\pdfoutput=1

\documentclass[11pt]{article}

\usepackage{style}
\usepackage{multirow}
\usepackage{booktabs}
\usepackage{adjustbox}

\usepackage{times}
\usepackage{latexsym}

\usepackage{graphicx}
\usepackage{mathtools} 
\usepackage{tikz}

\usepackage{hyperref}

\usepackage{amsmath}
\usepackage{amssymb}
\usepackage{mathtools}
\usepackage{amsthm}
\usepackage{xspace}
\usepackage{bm}
\usepackage{bbm}
\usepackage{enumitem}
\usepackage{float}
\usepackage{natbib}

\usepackage[T1]{fontenc}

\usepackage[utf8]{inputenc}

\usepackage{microtype}

\usepackage{inconsolata}

\theoremstyle{definition}

\theoremstyle{remark}

\newcommand{\mc}[1]{\ensuremath{\mathcal{#1}}\xspace}
\newcommand{\mb}[1]{\ensuremath{\mathbf{#1}}\xspace}
\newcommand{\tn}[1]{\ensuremath{\textnormal{#1}}\xspace}
\newcommand{\ol}[1]{\ensuremath{\overline{#1}}\xspace}

\DeclareMathAlphabet\mathbfcal{OMS}{cmsy}{b}{n}

\newcommand{\mbc}[1]{\ensuremath{\mathbfcal{#1}}\xspace}

\newcommand{\bU}{{\mb{U}}}

\newcommand{\fgl}{{{\sc FgLab}}\xspace}
\newcommand{\fgs}{{{\sc FgScr}}\xspace}
\newcommand{\pfgs}{{{\sc PFgScr}}\xspace}
\newcommand{\psl}{{{\sc PsLab}}\xspace}

\title{FRACTAL: Fine-Grained Scoring from Aggregate Text Labels}

\author{%
Yukti Makhija \\ Google Research India \\  {\tt yuktimakhija@google.com}
\and
Priyanka Agrawal \\ Google DeepMind \\ {\tt priyankagr@google.com}
\and
Rishi Saket \\ Google Research India \\ {\tt rishisaket@google.com}
\and
Aravindan Raghuveer \\ Google Research India \\ {\tt araghuveer@google.com}
}

\begin{document}
\maketitle
\begin{abstract}
Large language models (LLMs) are being increasingly tuned to power complex generation tasks such as  writing, fact-seeking, querying and reasoning. Traditionally, human or model feedback for evaluating and further tuning LLM performance has been provided at the response level, enabling faster and more cost-effective assessments. However, recent works (\cite{amplayo2022smart}, \cite{wu2023finegrained}) indicate that sentence-level labels may provide more accurate and interpretable feedback for LLM optimization. In this work, we introduce methods to disaggregate response-level labels into sentence-level (pseudo-)labels. Our approach leverages multiple instance learning (MIL) and learning from label proportions (LLP) techniques in conjunction with prior information (e.g., document-sentence cosine similarity) to train a specialized model for sentence-level scoring. We also employ techniques which use model predictions to pseudo-label the train-set at the sentence-level for model training to further improve performance. We conduct extensive evaluations of our methods across six datasets and four tasks: retrieval, question answering, summarization, and math reasoning. Our results demonstrate improved performance compared to multiple baselines across most of these tasks.  Our work is the first to develop response-level feedback to sentence-level scoring techniques, leveraging sentence-level prior information, along with comprehensive evaluations on multiple tasks as well as end-to-end finetuning evaluation showing performance comparable to a model trained on fine-grained human annotated labels. 
\end{abstract}

\section{Introduction} \label{sec:intro}

Large language models (LLMs) demonstrate a remarkable ability to generate text, seek facts, answer complex queries, and perform logical reasoning tasks.  Their progress is driven largely by a complex interplay of model architecture, training data, and tuning procedures.  The process of LLM refinement relies heavily on their evaluation and preference feedback, typically from humans or automated model-based scoring.
These feedback and scores are commonly used during various phases of LLMs development like model evaluation, reinforcement learning (RLHF/ RLAIF), bulk data distillation, and model hedging. However, such feedback has typically been taken at the response level, enabling efficient and cost-effective assessments of overall output quality.

An emerging body of research (\cite{amplayo2022smart}, \cite{lightman2023lets}) suggests that the sentence or step-level evaluation is more reliable and precise over response-level evaluation. Segment-level feedback promises improved accuracy by localizing strengths and weaknesses within a generated response. It further provides greater interpretability, allowing for more targeted LLM fine-tuning by highlighting the specific portions of a response that contribute to or detract from its overall quality. Moreover, collecting finer-grained human feedback is shown to result in considerably improved LLM training \cite{wu2023finegrained}. 

Even in situations where it is  feasible to directly collect fine-grained feedback, doing so for Side-by-Side (SxS) feedback could remain challenging and might also lead to significantly more expensive annotation process. 

To overcome this lack of fine-grained feedback, 
our work proposes methods to dis-aggregate response-level labels into sentence-level \emph{pseudo}-labels that accurately reflect the underlying quality distribution within a larger response. 

As in supervised training, the first component of our solution is a methodology to train a model on the response-labels to predict the scores (label probabilities) for instances. For this we leverage and build upon techniques from multiple instance learning (MIL) and learning from label proportions (LLP). These have been used to train predictive models on datasets partitioned into \emph{bags} or sets of \textit{instances} (e.g. feature-vectors or sentences). Each bag has an aggregated label i.e., bag-label which is thought to be derived from the (unknown) instance-labels of the bag via an aggregation function. In MIL, the aggregation is the ${\sf MAX}$ or ${\sf MIN}$ of binary or ordinal instance-labels  -- applicable to question-answering (relevance), summarization and math reasoning tasks -- while LLP, which models the bag-label as the ${\sf AVG}$ of the instance-labels, is applicable to retrieval tasks.
A standard technique in MIL and LLP is \emph{bag-loss} which uses some approximation of the aggregation function to compute the aggregated model prediction for each bag, and minimizes a loss between the bag-labels and the aggregated predictions, summed over all bags. While bag-loss is usually a strong baseline, it does not use any instance-level knowledge to guide the optimization. Such application-specific information can the modeled as a \emph{prior} distribution on the instance-labels. 
Our main technical contributions include designing priors -- based on document-sentence similarity and correlations -- for the various natural language tasks the we study. These priors  are used to  enhance bag-loss via an additional loss term, complementing the weak supervision provided by the bag-labels.

While the trained model predictions are probabilities for the instance-labels, we also develop \emph{pseudo-labeling} mechanisms to label instances using model predictions, while ensuring  consistency with the bag-labels. This provides a proxy for instance-label supervision and we demonstrate that training a model on these labels can improve the performance for downstream tasks.
\begin{figure*}[ht]
\centering
\includegraphics[width=\textwidth]{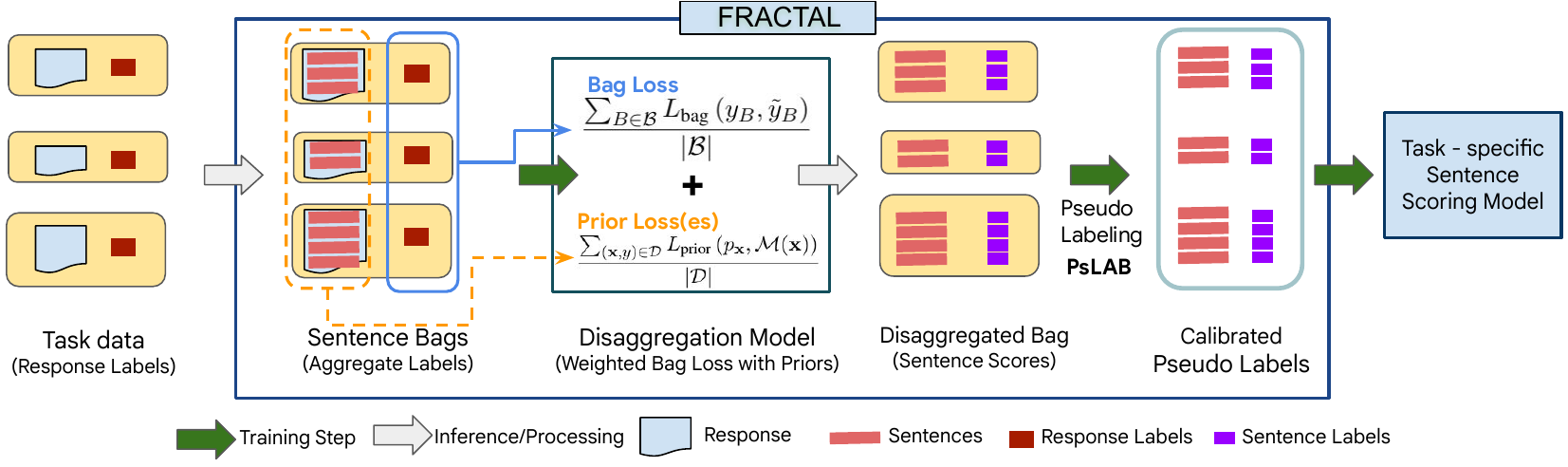}
\caption{Overview of our proposed method, FRACTAL. Input is a set of responses each with a response label. A response is a bag of sentences. The output is a model that can predict the score for each sentence in a response. The semantic meaning of a score depends on how the response label was defined.  FRACTAL consists of three key components a) Loss Function Design (Section~\ref{sec:bagloss}). b) Differentiable Approximations of Aggregation Functions (Section~\ref{sec:min_approx_main}) c) Max-Likelihood Pseudolabeling (Section~\ref{sec:pseudolab})  }
\label{fig:approach}
\end{figure*}

\noindent
Our contributions include:
\begin{enumerate}
    \item We study retrieval, question answering, summarization, and math reasoning tasks, and define various formulations for disaggregating response-level labels to sentence-level labels applicable to these tasks.
    \item We define the corresponding formulations of learning from response-level labels as MIL and LLP problems, as well as the sentence (instance) level priors for these tasks based on document-sentence similarity scores  and correlations between sentences. 
    \item We propose enhancements of the bag-loss method, by adding prior information as new loss loss terms, for training instance-level models.
    \item We develop pseudo-labeling strategies to transform instance-level model predictions into labels which are consistent with the response-level labels, allowing us to train the model on the derived pseudo-labels.
    \item We perform a comprehensive evaluation of our methods across a diverse set of six datasets spanning the studied tasks.  Our results demonstrate performance improvements over established baselines.
\end{enumerate}
We call our method for generating instance-level scores as FRACTAL, whose elements are illustrated in Figure \ref{fig:approach} and described in more detail in Sec. \ref{sec:our_techniques}.

\section{Related Work} \label{sec:previous_work}
 
\noindent
 {\bf Multiple Instance Learning (MIL).} Here the bag label is modeled as ${\sf MAX}$ or ${\sf MIN}$ of the its (unknown) instance-labels (typically labels are $\{0,1\}$-valued). Given a such a dataset of bags the goal is to train a model either to predict the bag-labels or in many cases the label of instances. This formulation was introduced to model drug activity detection in the work of \cite{DLL97} and was shown thereafter to have applicability in several other domains including drug discovery~\cite{ML97}, time series prediction~\cite{M98}, information retrieval~\cite{LY00} and the analysis of medical images~\cite{WYY15} and videos~\cite{BYB09,SDB13}. 
The work of \cite{ramon2000multi} proposed the bag-loss method using the log-sum exponential approximation to ${\sf MIN}$, and was followed by adaptations of  boosting and logistic regression using similar differentiable approximations~\cite{ZPV05,RC05}. More specialized MIL methods such as diverse-density (DD)~\cite{ML97} and and its EM-based variant, EM-DD~\cite{ZG01} have also been developed. More recent works have proposed deep learning methods for MIL based on convolutional and attention mechanisms~\cite{wu2015deep,ilse2018attention}.

\noindent
{\bf Learning from Label Proportions (LLP).} This is a variant of the above in which the bag-label is the average of the instance-labels, with the goal of training an instance-label predictor from bag data, and arises in the context of label privacy concerns~\cite{R10}, costly supervision~\cite{CHR} or lack of labeling instrumentation~\cite{DNRS}. As in the MIL case, the early works on LLP applied traditional supervised learning techniques~\cite{FK05,MCO07,R10}. Later works \cite{QSCL09, PNCR14} directly estimated model parameters from bag-labels while \cite{YLKJC13} proposed a specialized SVM for LLP.  Subsequently, methods using bag pre-processing \cite{SZ20,SRR} and training deep networks~\cite{KDFS15, LWQTS19}
have been developed. In particular, \cite{ardehaly2017co} proposed the bag-loss method for LLP, which  -- known as DLLP in the literature --  has been used used as a baseline in subsequent works. 

\noindent
{\bf MIL and LLP for NLP.} Applications such as sentiment analysis~\cite{pappas-popescu-belis-2014-explaining,angelidis-lapata-2018-multiple} and document modeling~\cite{pappas2017explicit} have previously admitted MIL techniques, while more recently \cite{liu-etal-2022-multiple} modeled offensive language detection as an MIL problem and proposed a mutual attention based mechanism. On the other hand, the applications of LLP are relatively sparser: \cite{ardehaly2016domain} applied it to domain adaptation for text data, while recent work~\cite{chauhan-etal-2023-learning} proposed a novel method improving on the baseline model training technique of \cite{ardehaly2017co} for text classification.

For both MIL and LLP, previous works have proposed pseudo-labeling based model training methods, in which the weak-supervision of bag-labels is used along with model predictions to derive \emph{pseudo-labels} which can be used to train or fine-tune models. For e.g. pseudo-labels are computed via regularization~\cite{WFCCH23,LWSQT21} or expectation-maximization~\cite{LGSKWDX20,BK22} techniques.

\section{Preliminaries} \label{sec:prelims}
Let $\mbc{X}$ be the underlying set of instances  and $\mc{Y}$ be the label-set which is typically $\{0,1\}$ or $\{0,1,\dots, L\}$ for binary or integer labels respectively.  
A dataset is a collection of labeled instances.

A bag $B$ is a subset of $\mbc{X}$ and  $y_B$ denotes its label which is thought to depend on the labels of the instances in $B$ via an \emph{aggregation function} ${\sf Agg}$ which maps tuples with elements from $\mc{Y}$ to a bag label-set $\ol{\mc{Y}}$ which is the real-segment spanned by $\mc{Y}$ i.e., either $[0,1]$ or $[0,L]$. Specifically, if $B = \{\bx_1, \dots, \bx_k\}$ and $y_i$ is the label of $\bx_i$ ($i \in [k]$), then $y_B = {\sf Agg}\left(\bx_1, \dots, \bx_k\right)$. Typically, ${\sf AGG}$ is either ${\sf MIN}$, ${\sf MAX}$ or ${\sf AVG}$.

We consider \textit{prior information} about the labels on individual instances, for e.g. through some unsupervised modeling or side information. For each $\bx$ in the dataset, its prior $p_{\bx}$ is a soft label in $[0,1]$ or $[0, L]$. Another class of priors is given by $p_{\bx\bz}$ for pairs $(\bx, \bz)$ of instances. 

Some applications provide \emph{preference} bag-labels which encode comparisons between pairs of bags. Specifically, for a pair of bags $(B_1, B_2)$ the preference bag-label is $y_{B_1B_2}$ which is $1$ if $y_{B_1} > y_{B_2}$, and $-1$ if $y_{B_1} \leq y_{B_2}$. 

\medskip

\noindent
{\bf Modeling Task.} Given as input a collection $\mc{B}$ of pairwise-disjoint (i.e., non-overlapping) bags along with their bag-labels, possibly along with  the priors $\{p_{\bx}\}$ or $\{p_{\bx\bz}\}$, the goal is to output a model predicting a score for each instance in $\mbc{X}$. In the preference evaluation, we evaluate the model in terms of the accuracy of the preference labels assigned by the model on a test set of bags.

\section{Our Techniques} \label{sec:our_techniques}
We  present the the components of the FRACTAL method along with the BagLoss baseline approach.  
\subsection{Bag-loss with priors} \label{sec:bagloss}
Let us denote by ${\sf probAGG}$ some differentiabale approximation of ${\sf AGG}$ to soft-labels $\ol{Y}$. 
The BagLoss method optimizes the following loss, for a  model $\mc{M}$:
\begin{align}
 L_{\tn{totbag}}(\mc{B}, \mc{M}) := \frac{\sum_{B \in \mc{B}}L_{\tn{bag}}\left(y_B, \tilde{y}_B\right)}{|\mc{B}|} \label{eqn:bagloss}
\end{align}
where $\tilde{y}_B = {\sf probAGG}\left(\left(\mc{M}(\bx)\right)_{\bx \in B}\right)$ is the aggregate prediction of bag $B$, $L_{\tn{bag}}$ is some loss function.  
In our bag loss with priors, PriorsBagLoss method, we have an additional loss which incorporates the priors. For priors $\{p_{\bx}\}$ we have:
\begin{equation}
    L_{\tn{totprior1}}(\mbc{X}, \mc{M}) := \frac{\sum_{\bx \in \mbc{X}}L_{\tn{prior}}\left(p_{\bx}, \mc{M}(\bx) \right)}{|\mc{\mbc{X}}|} \label{eqn:priorloss}
\end{equation}
while for priors $\{p_{\bx\bz}\}$ defined over pairs of instances:
\begin{align}
    &L_{\tn{totprior2}}(\mbc{X}, \mc{M}) \nonumber \\ & := \frac{\sum_{(\bx,\bz) \in \mbc{X}^2}L_{\tn{prior}}\left(p_{\bx\bz}, \mc{M}(\bx)\mc{M}(\bz) \right)}{|\mbc{X}|^2} \label{eqn:priorloss-2}
\end{align}
where $L_{\tn{prior}}$ is a loss. The total loss is a combination of bag and prior losses:\\
    $L_{\tn{tot}} = \lambda L_{\tn{totbag}} + \lambda_1L_{\tn{totprior1}} +  \lambda_2L_{\tn{totprior2}}$,\\ for some $\lambda,\lambda_1,\lambda_2 \in [0,1]$, s.t. $\lambda + \lambda_1 + \lambda_2 = 1$.

\noindent
{\bf Minibatch based model training.} For a given batch size $q$, learning rate and optimizer as well as hyperparameter $\lambda \in [0,1]$, we train the predictor model $\mc{M}$  by doing the following for $N$ epochs and $K$ steps per epoch:
\begin{enumerate}[noitemsep,nolistsep]
    \item Sample a minibatch $S$ of $q$ bags $\mc{B}_S \subseteq \mc{B}$.
    \item Using current model predictions, compute $L_{\tn{totbag}}$ and $L_{\tn{totprior}}$ restricted only to the bags and instances in $S$, and compute $L_{\tn{tot}}$.
    \item Using the  required gradients  from \eqref{eqn:bagloss}, \eqref{eqn:priorloss} and \eqref{eqn:priorloss-2} along with the optimizer and learning rate, update the weights of the model $\mc{M}$.
\end{enumerate}

\subsubsection{Preference based bag-loss with priors}\label{sec:prefbagloss}
The PrefBagLoss approach is similar to those in the previous subsection, where instead of $L_{\tn{totbag}}$ we have a preference based loss for the pairs of bags $S$ for which preference labels are available. For a pair of bags $(B_1, B_2)$ with $y_{B_1B_2}$ be the preference-label let $\tilde{y}_{B_1}$ and $\tilde{y}_{B_2}$ be positive real-valued aggregate predictions of $B_1$ and $B_2$ respectively. Using the \cite{Bradley-Terry} model we have the loss:
\begin{equation}
    L_{\tn{pref}}(B_1, B_2, y_{B_1B_2}) := y_{B_1B_2}\log \frac{y_{B_2}}{y_{B_1}}
\end{equation}                              
which is averaged over all pairs in $S$ to obtain $L_{\tn{totpref}}$ which is minimized.

The minibatch training now samples pairs of bags and computes $L_{\tn{totpref}}$ restricted to the sampled pairs. In the priors based extension, PriorsPrefBagLoss, the $L_{\tn{totprior}}$ loss remains the same, over all the instances in the minibatch.

\subsection{Approximations to {\sf MIN}} 
\label{sec:min_approx_main}
In the binary-label case, the standard baseline is ${\sf Mult}$ which is just the product of the probabilities. 
We employ the in built TensorFlow (TF) approximation ${\sf tf\_reduce\_min}$ in our experiments (See Appendix \ref{sec:minappox} for more details and ablations with other approximations).  Note that ${\sf MAX}$ can be derived from ${\sf MIN}$ applied to flipped variables in the binary case. For integer labels we use the respective TF approximations for ${\sf MIN}$ and ${\sf MAX}$.

\subsection{Max-Likelihood Pseudo-labeling and Model Training}\label{sec:pseudolab}
Our PsLab pseudo-labeling method uses the prediction of the model $\mc{M}$ trained as per the techniques described above, to output the max-likelihood instance-labels for each bag, consistent with the bag-label. We apply this to the case of $\{0,1\}$-labels and model predictions being probabilities, and the ${\sf MIN}$ aggregation (the ${\sf MAX}$ case is analogous). 
For a given bag, consider the distribution in which each $\bx$ in the bag is independently $1$ with probability $\mc{M}(\bx)$ and $0$ otherwise. If the bag-label is $1$ then PsLab outputs $1$ for each instance in the bag. 
When the bag-label is $0$, then PsLab outputs the maximum likelihood valid configuration of labels i.e.,  with at least one $0$-label. This is can be efficiently done via the following algorithm:
\begin{enumerate}[nolistsep,noitemsep]
    \item Compute the labeling $\Gamma: B \to \{0,1\}$ where $\Gamma(\bx) = 1$ if $\mc{M}(\bx) > 1$ and $0$ otherwise.
    \item If $\Gamma(\bx) = 1$ for all $\bx \in B$, then modify $\Gamma$ by flipping the label of $\bz\in B$ which has the minimum value of $\mc{M}(\bz)$. 
\end{enumerate}

\smallskip
\noindent
{\bf Model Training on Pseudo-labels.} After computing the pseudo-labels on the training bags, we now have a train-set with labeled instances. The model retrained on this dataset is evaluated for comparative performance.

\section{Tasks and Datasets}
\label{sec:datasets}

Tables \ref{tab:my_label} and \ref{tab:my_label-2} provide concise descriptions of the tasks and datasets that we consider. In the rest of this section we provide more details on various aspects of these datasets and tasks.

\begin{table*}[]
    \begin{small}
    \centering
    \begin{adjustbox}{width=\linewidth}
    \begin{tabular}{p{4cm}|p{3cm}p{4cm}p{6cm}p{3cm}}
    \toprule
     \textbf{Dataset (Input for training)} & \textbf{Bags} & \textbf{Instances} & \textbf{Task} & \textbf{Prior}\\
     \midrule 
      \textbf{QA-Feedback} (Question, Knowledge Passage, Pair of Responses, Preference Label) & Both responses are treated as separate bags. & Sentences in a response + question + Knowledge Passages  & Learn sentence-level scores for relevance using only the preference bag-label (indicates which response is better). The aggregation function used is  {\sf AVG}  &  knowledge passage-sentence cosine sim., corr. b/w sentences.\\
      \midrule
     \textbf{FirA} (Paragraph, Query, Relevance Score) & Paragraph & sentence of the paragraph + query & Learn $\{0,1,2,3,4\}$-valued relevance score for each sentence wrt query. The aggregation function used is {\sf MAX} & query-sentence cosine similarity \\
     \midrule
      \textbf{MultiSpanQA} (Context, Question, Label (answer present in context) & Context & Sentences of the context + Questions & Identify sentences of the context which contain the answer to the question. We use {\sf MAX} for this binary classification setup. & query-sentence cosine sim., corr. b/w sentences\\
      \midrule
      \textbf{AquaMuSe} (Documents, Query, Summary, Entailment Label) & Summary & Sentence of the summary  + documents + query & Given a query, document and bag-level binary entailment label, determine the non-entailed sentences in a summary. {\sf MIN} is the aggregate function used in this setup. & doc-sentence cosine sim., corr. b/w sentences \\
      \midrule
    \textbf{WikiCatSum} (Documents, Summary, Entailment Label) & Summary & Sentence of a summary  + documents & Given a document and bag-level binary entailment label, determine the non-entailed sentences in a summary. {\sf MIN} is the aggregate function used in this setup. & doc-sentence cosine sim., corr. b/w sentences \\
    \midrule
    \textbf{PRM800K} (MATH Problem, step-wise solution, Label(correctness) & Solution to the MATH problem & Step of the solution + question & Using the binary aggregate label indicating the correctness of the solution, identify all incorrect steps in the solution. & question-step cosine sim., corr. b/w steps \\
    \bottomrule
    \end{tabular}
    \end{adjustbox}
    \caption{Summary of the bags, labels, instances, annotations and priors for each dataset.}
    \label{tab:my_label}
    \end{small}
\end{table*}

\begin{table}[htb]
    \centering
    \begin{adjustbox}{width=0.5\columnwidth}
    \begin{tabular}{cc|cccccccc}
    \toprule
     \textbf{Task} & \textbf{Dataset} & \textbf{Objective}  & \textbf{Splits (Train|Test)} \\
     \midrule 
      Long-form QA & QA-Feedback  & Preference &  13.5k | 3k\\
       Retrieval & FirA  &Regression&  18k | 4k\\
       Retrieval & MultiSpanQA  &Classification &  5k | 650 \\
       Summarization & AquaMuSe  & Classification & 3k | 500 \\
       Summarization & WikiCatSum  & Classification & 45k | 1.5k\\
       Math Reasoning & PRM800K & Classification & 1k | 100\\
    \bottomrule
    \end{tabular}
    \end{adjustbox}
    \caption{Summary of the setup used for each dataset}
    \label{tab:my_label-2}
\end{table}

\paragraph{\textbf{Long-form Question Answering.}}

We use QA-Feedback dataset, an SxS preference dataset collected and released by \cite{wu2023finegrained}. These are human provided preferences  on pairs of model generated responses for input questions and relevant passages from ASQA~\cite{stelmakh2023asqa}, an ambiguous factoid QA dataset. The data further contains segment-level annotations, and we make use of ``irrelevance, repetition, or incoherence'' category for sentence-level evaluation. The responses are the bags and the preferences are the preference bag-labels.The aggregation function used for the bag-level loss term is {\sf AVG}, and the priors integrated into the loss function include the cosine similarity between knowledge passages and each sentence of the response. Specifically:
\begin{equation}
    \tn{P1}(\bx) := {\sf cosprior}(\bx) = \frac{1}{2}\left(1 + \frac{\langle \bx, \bU\rangle}{\|\bx\|_2\|\bU\|_2}\right) \label{eqn:cosprior}
\end{equation}
where $\bx$ and $\bU$ are (embeddings of) a sentence and the relevant passage, and we scale cosine-similarity so that its value is in $[0,1]$. 
Additionally, we incorporate the Pearson's correlation between a pair of sentences as a prior:
\begin{equation}
   \tn{P2}(\bx,\bz) := {\sf corrprior}(\bx,\bz) = \frac{1}{2}\left(1 + \rho_{\bx\bz}\right) \label{eqn:corrprior}
\end{equation}
where $\rho_{\bx\bz}$ is the Pearson's correlation between sentence embeddings $\bx$ and $\bz$.

\paragraph{\textbf{Retrieval.}} We use two datasets for retrieval tasks: MultiSpanQA and FiRA. MultiSpanQA~\cite{li-etal-2022-multispanqa} is aimed at questions that necessitate identifying multiple discontinuous text spans from a given passage to compose the complete answer. This dataset consists of question-context pairs, with annotated answer spans for the train and validation splits. We randomly select 25\% of the train-split as the test-split. Context is treated as a bag, with its instances being sentences which labeled $1$ if they overlap with annotated spans, and $0$ otherwise. The {\sf MAX} aggregation is used to indicate the presence an answer in the context. All MultiSpanQA samples contain answers to questions, resulting in all positive bags. To balance the dataset, negative bags are created for half of the samples by extracting context chunks without answers. Thus, both the instance and bag labels are $\{0,1\}$-valued.

The FiRA dataset~\cite{Hofstaetter2020_fira} comprises word-level relevance annotations using $\{0,\dots,4\}$-valued labels for the relevance of each word in a paragraph to a query. We compute the word-level average across annotators and then the maximum across all words in a sentence to derive sentence-level scores. Similar to the previous setup, we treat the paragraph as a bag, its sentences as instances, and employ {\sf MAX} as the aggregation function. The instance and bag-level scores range from $0$ to $4$, with the goal of optimizing a regression loss.

For both datasets, we integrate a correlation prior between sentence pairs and a cosine-similarity prior (see \eqref{eqn:cosprior}, \eqref{eqn:corrprior}) between the query and each sentence of the context. 
\paragraph{\textbf{Summarization.}} We utilize two datasets: WikiCatSum and AquaMuSe. The WikiCatSum dataset \cite{perez-beltrachini-etal-2019-wikicatsum} is specifically designed for multi-document summarization tasks, focusing on generating Wikipedia-style lead sections for entities within three domains: Companies, Films, and Animals out of which  we focus on the Films and Animals domains.
On the other hand, the AquaMuSe dataset \cite{kulkarni2020aquamuse} is tailored for multi-document, question-focused summarization.

We adopt the binary entailment metric for this task. The reference summaries already provided in these two datasets serve as the \textit{entailed} summaries\footnote{In this work, we do not filter any noise present in the existing data splits.}. Each sentence in these summaries is considered positively entailed. To generate \textit{non-entailed} summaries, we synthesize negatives by employing various manipulations, similar to \cite{Yin_2021docnli}. Firstly, we perturb the reference summary through sentence replacement. This involves randomly selecting $k$ sentences, where $k$ is less than the total sentences in the summary, and iteratively feeding their left context to an ULM to predict the next sentence. The predicted sentence is then used to replace the selected one. Additionally, we explore the standard word replacement technique, which randomly masks $k$ words whose POS tags are among proper nouns, numbers, and verbs, to introduce factual errors in the summaries. The masked words are then predicted using BERT. The number of replaced sentences and words is randomly selected for each sample. The perturbed sentences within the summary are considered non-entailed, while the remaining unchanged sentences are deemed entailed. Thus, the sentence as well as bag labels are $\{0,1\}$-valued with $1$ indicating entailed and $0$ non-entailed, with ${\sf MIN}$ as the aggregation function. Examples of entailed and non-entailed summaries are provided in Appendix \ref{sec:example_summary}.
As in previous tasks, we incorporate sentence-document cosine similarity and sentence correlation priors into our methods. Additionally, we experiment with NLI entailment scores \cite{Honovich2022TRUERF} as priors for this task. 

\paragraph{\textbf{Math Reasoning}} We utilize Phase 1 of the PRM800K dataset \cite{lightman2023lets} releasing step-level annotations for model-generated solutions to MATH problems \cite{hendrycks2021math}. The task at hand is to identify all the incorrect steps in the solution. 

\begin{table}[htb]
    \centering
    \begin{adjustbox}{width=0.5\linewidth}
    \scriptsize
    \begin{tabular}{l|cccccc}
    \toprule
    Method & FirA & MSQA & QA-FB & AqMse & WikiCS & PRM \\
    \midrule
    
    cos-sim & - & 0.455 & 0.535 & 0.632 & 0.408 & 0.42 \\ 
        Resp-level & 0.319 & 0.583 & 0.491 & 0.697 & - & 0.528 \\ 
        BagLoss & 0.304 & 0.661 & 0.509 & 0.751 & 0.477 & 0.569 \\ 
        \textbf{FRACTAL (ours)} & 0.294 & 0.693 & 0.532 & 0.814 & 0.645 & 0.597 \\ 
        Supervised & 0.283 & 0.729 & 0.651 & 0.876 & 0.837 & 0.613 \\ 
    \bottomrule
    \end{tabular}
    \end{adjustbox}
    \caption{We compare our method against several baselines across all datasets. The columns left to right are for the FiRA, MultiSpanQA, QA-Feedback, WikiCatSum, AquaMuse and PRM800K datasets. For the FiRA dataset, we report MAE, while for the others we report AUC-ROC.     }
    \label{tab:overview_table}
\end{table}

\section{Experiments}
We evaluate FRACTAL along with baseline techniques (listed below) on the tasks and datasets describes in Sec. \ref{sec:datasets}.
\paragraph{\textbf{Off the shelf Baselines.}} The following methods directly score the sentences: \\
\emph{Semantic Similarity:} This uses similarity of individual sentences from the response with the input context to estimate their relevance for the task. For this, we use the cosine similarity (see \eqref{eqn:cosprior}) between the corresponding embeddings. \\
\emph{Entailment Scorer:} For summarization and relevance tasks, we also compute entailment using the NLI scorer from TRUE paper \cite{Honovich2022TRUERF}. This is a T5x-11B model \cite{raffel2023exploring} finetuned on several NLI datasets. 

\noindent
\paragraph{\textbf{Trainable Baselines.}} These use the bag or instance labels to train models. \\
{\emph{BagLoss}}: This uses BagLoss on bag-labels (or PrefBagLoss in case of preference bag-labels) described in Sec. \ref{sec:our_techniques}.\\
\emph{Response-level:} This uses embeddings for responses and the corresponding response-labels for model training while inference is on sentences.\\
\emph{Supervised:} This trains directly on instance-labels to provide an upper baseline for comparison.
\smallskip
\noindent
{\bf FRACTAL}: As described in Sec. \ref{sec:our_techniques}, this involves PriorsBagLoss (or PriorsPrefBagLoss) based model training using bag-labels (or preference bag-labels) as well priors, TensorFlow approximations to the the ${\sf MIN}$ or ${\sf MAX}$ functions. The PsLab pseudo-labeling uses the best performing model (trained using prior augmented bag loss) predictions to pseudo-label the train-set followed by another model training on these resultant instance (pseudo)-labeled train-set. Note that (i) when we only have preference bag-labels, or (ii) when the model prediction is a value in $[0,L]$ with $L > 1$ so that label-probabilities are not available, PsLab is not applicable, and FRACTAL provides the model trained on our prior augmented bag loss methods.
The $L_{\tn{bag}}$ and $L_{\tn{prior}}$ losses are taken to be cross-entropy except for mse for the regression objective on the FiRA dataset for which the priors are also appropriately scaled.

\begin{table*}[h!]
\centering
\begin{adjustbox}{width=0.4\linewidth}
\begin{tabular}{@{}lrrr@{}}
\toprule
\textbf{Method} &
\multicolumn{1}{l}{\textbf{AUC-ROC}} &
\multicolumn{1}{l}{\textbf{AUC-PR}} &
\textbf{Accuracy}  \\ 
\midrule
\textbf{\emph{MultiSpanQA}} & \\
Supervised     & 0.729 ± 0.016 & 0.354 ± 0.008 & 0.861 ± 0.046\\
Cosine Similarity &  0.455 &  0.135 & 0.851 \\
NLI & 0.631         & 0.366         & \textbf{0.859} \\
Response-level Model     & 0.583 ± 0.187 & 0.217 ± 0.094 & 0.852 ± 0.003\\
BagLoss                     & 0.661 ± 0.092 & 0.309 ± 0.127 & 0.852 ± 0.133 \\  
\underline{FRACTAL Methods} \\
PriorBagLoss(0.2, 0)        & 0.669 ± 0.063 & 0.311 ± 0.059 & 0.838 ± 0.071  \\
PriorBagLoss(0.1, 0.2) & 0.625 ± 0.07  & 0.271 ± 0.039 & 0.851 ± 0.021  \\
\psl &
\textbf{0.693 ± 0.115} &
\textbf{0.326 ± 0.071} &
0.842 ± 0.052 \\ 

\midrule
\textbf{\emph{QA Preference Feedback}} & \\
Supervised         & 0.651 ± 0.009 & 0.609 ± 0.007  & 0.688 ± 0.011  \\
Cosine Similarity           & 0.535 & 0.526  & 0.483  \\
Response-level Model        & 0.491 ± 0.008 & 0.4643 ± 0.007 & 0.453 ± 0.015\\
PrefBagLoss                     & 0.509 ± 0.005 & 0.526 ± 0.002 & 0.515 ± 0.009\\
\underline{FRACTAL Methods} \\
PriorPrefBagLoss(0.2, 0)        & 0.516 ±  0.003 & 0.494 ± 0.003 & 0.509 ± 0.007\\
PriorPrefBagLoss(0, 0.4)        & 0.528 ± 0.003 & 0.519 ± 0.002  & \textbf{0.530 ± 0.006}  \\
PriorPrefBagLoss(0.2, 0.5) & \textbf{0.532 ± 0.004} & \textbf{0.526 ± 0.005} & 0.521 ± 0.008 \\

\midrule 
\textbf{\emph{WikiCatSum}} & \\
Supervised & 0.837 ± 0.062 & 0.894 ± 0.085 & 0.718 ± 0.085  \\
Cosine Similarity & 0.408 & 0.829 & 0.362 \\
NLI & 0.639 & 0.817 & 0.648  \\
BagLoss & 0.477 ± 0.093 & 0.831 ± 0.052 & 0.562 ± 0.047 \\
\underline{FRACTAL Methods} \\
PriorBagLoss(0.2, 0.1, 0) & 0.641 ± 0.028 & 0.879 ± 0.013 & 0.651 ± 0.017 \\
PriorBagLoss(0, 0, 0.4) & 0.642 & \textbf{0.885} & 0.653 \\
\psl & \textbf{0.645 ± 0.038} & 0.879 ± 0.057 & \textbf{0.663 ± 0.091} \\

\midrule
\textbf{\emph{AquaMuSe}} & \\
Supervised   & 0.876 ± 0.007 & 0.926 ± 0.002 & 0.866 ± 0.008\\
Cosine Similarity    & 0.632    & 0.763 &  0.649\\
NLI & 0.793 & 0.889 & 0.824 \\
Response-level Model & 0.696 ± 0.011 & 0.775 ± 0.009 & 0.675 ± 0.015 \\
BagLoss   & 0.747 ± 0.007 & 0.824 ± 0.005 &  0.779 ± 0.01\\
\underline{FRACTAL Methods} \\
PriorBagLoss(0.2, 0.1) & 0.789 ± 0.004 & 0.871 ± 0.006 & 0.831 ± 0.007 \\
\psl & \textbf{0.814 ± 0.006}  & \textbf{0.898 ± 0.007} & \textbf{0.834 ± 0.01} \\

\midrule
\textbf{\emph{PRM800K}} & \\
Supervised & 0.613 ± 0.028 & 0.928 ± 0.021 & 0.709 ± 0.051  \\
Cosine Similarity    & 0.420         & 0.873         & 0.496        \\
Response-level Model & 0.528 ± 0.145 & 0.895 ± 0.037 & 0.521 ± 0.082 \\
BagLoss                 & 0.569 ± 0.019 & 0.924 ± 0.011 & \textbf{0.688 ± 0.071}  \\
\underline{FRACTAL Methods} \\
PriorBagLoss(0.5, 0.5)    & 0.582 ± 0.034 & 0.927 ± 0.009 & 0.534 ± 0.044 \\
PriorBagLoss(0, 0.1)   & 0.579 ± 0.052 & 0.925 ± 0.017 & 0.603 ± 0.038  \\
PriorBagLoss(0.1, 0.1) & 0.580 ± 0.068    & 0.926 ± 0.024   & 0.563 ± 0.049   \\
\psl & \textbf{0.597 ± 0.093} & \textbf{0.927 ± 0.004} & 0.578 ± 0.063 \\ 
\bottomrule
\end{tabular}
\end{adjustbox}
\caption{Instance-level Evaluation on MultiSpanQA, QA Preference Feedback, WikiCatSum, AquaMuSe and PRM800K Datasets}
\label{tab:all_except_fira_results_test_split}
\end{table*}

\begin{table}[htb]
\centering
\scriptsize
\begin{tabular}{llcc}
\toprule
 \textbf{Method} & \multicolumn{1}{l}{\textbf{MAE}} & \multicolumn{1}{l}{\textbf{MSE}} \\
\midrule
Supervised & 0.283 ± 0.072 & 0.141 ± 0.088 \\
Response-level Model & 0.319 ± 0.047 & 0.186 ± 0.098 \\
BagLoss & 0.304 ± 0.007 & 0.163 ± 0.002\\
\underline{FRACTAL Methods} \\
PriorBagLoss(0.3, 0) & 0.298 ± 0.002 & 0.157 ± 0.004\\
PriorBagLoss(0.2, 0.2) & \textbf{0.294 ± 0.003} & \textbf{0.155 ± 0.001}\\
 \bottomrule
\end{tabular}
\caption{Instance-level Evaluation on FiRA Dataset}
\label{tab:fira_results_test_split}
\end{table}

\begin{table}[]
    \centering
    \scriptsize
    \begin{tabular}{l|c}
        \toprule
       \textbf{Method}
         & \textbf{ROUGE} \\
         \midrule
        SFT + Preference RLHF & 48.96 \\
        SFT + FineGrained RLHF (Human Annotation) & 49.12 \\
        SFT + FineGrained RLHF (FRACTAL Prediction) & 49.04 \\
        \bottomrule
    \end{tabular}
    \caption{We perform fine-grained RLHF on the QA-Feedback dataset using the framework provided by \citet{wu2023finegrained} by replacing the Human Annotations  with relevance label predictions from a model trained only on preference labels.}
    \label{tab:RLHF}
\end{table}

\subsection{Model Training Setup}
We use the same model architecture across all tasks:  a Sentence-T5 Large encoder to generate embeddings for text components, followed by a 2-hidden layer MLP with 73728 parameters for predicting sentence-level scores. To handle lengthy documents exceeding 2000 tokens in MultiSpanQA, WikiCatSum, and AquaMuSe datasets, we partition documents into 1000-token paragraphs which are encoded separately to improve embedding quality. Subsequently, attention weights representing importance are learnt for each document split, and the document embedding is obtained through a weighted sum of individual split embeddings.  We report results averaged over 3 randomly seeded trials. 
We conduct grid search hyperparameter tuning to identify optimal parameter configurations, including learning rates, weights of prior terms integrated into the loss function, and batch sizes. The list of optimal hyperparameters for each dataset is provided in Appendix \ref{sec:hyper}.

\subsection{Results and Discussion}
Tables \ref{tab:all_except_fira_results_test_split} and \ref{tab:fira_results_test_split} provide the detailed evaluations of the baselines and our methods on the test data. In the tables, PriorsBagLoss$(\lambda_1,\lambda_2)$ and PriorsPrefBagLoss$(\lambda_1,\lambda_2)$ denote the instantiation of these methods with weights $\lambda_1$ and $\lambda_2$ for the losses corresponding to priors P1 and P2 respectively (see \eqref{eqn:cosprior}, \eqref{eqn:corrprior} and Sec. \ref{sec:bagloss}, \ref{sec:prefbagloss}).  For the WikiCatSum dataset, we incorporate NLI entailment scores as a prior, assigning a weight of $\lambda_3$ for the corresponding loss term. PsLab denotes the performance of the model trained after pseudo-labeling the train-set using the best performing prior augmented bag loss. Table \ref{tab:overview_table} summarizes the performance of the different baselines listed above and our proposed approach, FRACTAL on the tasks/datasets in Sec. \ref{sec:datasets}.

\noindent
\textbf{FRACTAL renders more precise sentence-level scores.}  
From Table \ref{tab:overview_table},
 we observe a consistent improvement in the sentence scoring over the BagLoss as well as the Response-level baseline across all these datasets in terms of AUC-ROC, bridging the performance gap between them and the best approach supervised (sentence-level trained) model. The use of priors along with pseudo-labeling based model training allows for an improved estimation of task specific score for sentences. It is interesting to note that BagLoss outperforms the Response-level baseline, suggesting that introduction of aggregate loss based methods to estimate sentence-level scores is itself useful. Response-level model trained with large response doesn't generalize well to smaller sentences.

\smallskip
\noindent
\textbf{Leveraging Prior improves Feedback Disaggregation.} Among the key ideas of FRACTAL is to augment BagLoss with cosine-similarity and correlation priors at the sentence-level (see \eqref{eqn:cosprior} and \eqref{eqn:corrprior}). Using a weighted combination of combination of BagLoss and the prior loss terms, provides substantially improved performance over either using just the BagLoss, or the cosine-similarity baseline. In effect, our proposed combination performs better than either of its constituents. We hypothesize that the priors provide sentence-level insight which complements the aggregate label based optimization of BagLoss. 
While performance of FRACTAL is sensitive to the weights of each loss component, during our hyperparameter sweep, we find that the model tends to select an upweighted BagLoss term for most datasets and metrics except for QA Preference Feedback.

\smallskip
\noindent
\textbf{Calibrated Pseudo-Labels are more helpful.}
 An interesting question is whether the disaggregated scores should eventually be calibrated such that their aggregate matches the response scores for training the final model, as done by our PsLab method in FRACTAL as described in  \ref{sec:pseudolab}. The ablations on PsLab are highlighted for Retrieval, Summarization, and Math Reasoning in Table \ref{tab:all_except_fira_results_test_split}. For all tasks, PsLab results in a more accurate model compared to both the baselines and the models trained on prior augmented BagLoss.

\smallskip
\noindent
\textbf{Downstream Tasks benefit from sentence-level scoring.} 
 \citet{wu2023finegrained} showed that using fine grained human labels per segment level across 3 categories irrelevant, untruthful, completeness can help improve performance. In the above setup, we replace the fine grained human labels of relevance with the FRACTAL predictions (Refer  Long-Form Question Answering task in Section~\ref{sec:datasets}). We observe in Table~\ref{tab:RLHF} that FRACTAL (Row 3) has comparable performance to Fine Grained RLHF with human labels (Row 2) and   FRACTAL predictions help improve performance over using only preference labels (Row 1). 

\smallskip
\noindent
\textbf{FRACTAL generalizes well across task objectives and aggregation functions} As mentioned in \ref{sec:datasets}, we experiment with a different task setups to thoroughly investigate the applicability of our approach to various LLM capabilities. As highlighted earlier in the overview and tasks specific Tables \ref{tab:overview_table}-\ref{tab:fira_results_test_split}, we find FRACTAL to be the best performing method. The task specific discussion of experimental results is deferred to Appendix \ref{sec:dataset_results_discussion}.

\section{Conclusions}
Our works casts the problem of deriving sentence-level scores from response-labels for complex text generation tasks as that of learning from aggregated labels in the MIL and LLP frameworks. We propose a novel method FRACTAL, which augments bag-loss using instance level priors to train predictor models, along with a pseudo-labeling technique for improved model training. Extensive evaluations of FRACTAL along with vanilla bag-loss and response-level model training baselines, as well as off-the-shelf scorers demonstrate substantial performance gains from FRACTAL models on six datasets spanning four tasks: retrieval, question answering, summarization, and math reasoning.
\section{Limitations}
Our method for label calibration and pseudo-labeling works well in classification tasks, leading to better performance. However, it's not as effective in regression tasks due to lack of label-specific model predictions (probabilities). Also, applying this technique becomes difficult when dealing with preference feedback. 
\section{Ethical Considerations}
We propose techniques to transform via model training methods, a response-level score into sentence-level scores. While such response-level scores are commonly obtained by human annotators, we have not conducted any human annotations and our evaluations are on publicly available datasets. Our method produces artificial labels for downstream tasks and does not modify in anyway the original response scores or attempt to associate them to any individual(s). 

\bibliographystyle{plainnat}
\bibliography{references}

\begin{thebibliography}{50}
\providecommand{\natexlab}[1]{#1}
\providecommand{\url}[1]{\texttt{#1}}
\expandafter\ifx\csname urlstyle\endcsname\relax
  \providecommand{\doi}[1]{doi: #1}\else
  \providecommand{\doi}{doi: \begingroup \urlstyle{rm}\Url}\fi

\bibitem[Amplayo et~al.(2022)Amplayo, Liu, Zhao, and Narayan]{amplayo2022smart}
Reinald~Kim Amplayo, Peter~J. Liu, Yao Zhao, and Shashi Narayan.
\newblock Smart: Sentences as basic units for text evaluation, 2022.

\bibitem[Angelidis and Lapata(2018)]{angelidis-lapata-2018-multiple}
Stefanos Angelidis and Mirella Lapata.
\newblock Multiple instance learning networks for fine-grained sentiment
  analysis.
\newblock \emph{Transactions of the Association for Computational Linguistics},
  6:\penalty0 17--31, 2018.
\newblock \doi{10.1162/tacl_a_00002}.
\newblock URL \url{https://aclanthology.org/Q18-1002}.

\bibitem[Ardehaly and Culotta(2016)]{ardehaly2016domain}
Ehsan~Mohammady Ardehaly and Aron Culotta.
\newblock Domain adaptation for learning from label proportions using
  self-training.
\newblock In \emph{IJCAI}, pages 3670--3676, 2016.

\bibitem[Ardehaly and Culotta(2017)]{ardehaly2017co}
Ehsan~Mohammady Ardehaly and Aron Culotta.
\newblock Co-training for demographic classification using deep learning from
  label proportions.
\newblock In \emph{2017 IEEE International Conference on Data Mining Workshops
  (ICDMW)}, pages 1017--1024. IEEE, 2017.

\bibitem[Babenko(2008)]{babenko2008multiple}
Boris Babenko.
\newblock Multiple instance learning: algorithms and applications.
\newblock \emph{View Article PubMed/NCBI Google Scholar}, 19, 2008.

\bibitem[Babenko et~al.(2009)Babenko, Yang, and Belongie]{BYB09}
Boris Babenko, Ming-Hsuan Yang, and Serge Belongie.
\newblock Visual tracking with online multiple instance learning.
\newblock In \emph{2009 IEEE Conference on Computer Vision and Pattern
  Recognition}, pages 983--990, 2009.
\newblock \doi{10.1109/CVPR.2009.5206737}.

\bibitem[Barucic and Kybic(2022)]{BK22}
Denis Barucic and Jan Kybic.
\newblock Fast learning from label proportions with small bags.
\newblock In \emph{Proc. {IEEE} {ICIP}}, pages 3156--3160, 2022.

\bibitem[Bradley and Terry(1952)]{Bradley-Terry}
Ralph~Allan Bradley and Milton~E. Terry.
\newblock Rank analysis of incomplete block designs: I. the method of paired
  comparisons.
\newblock \emph{Biometrika}, 39\penalty0 (3/4):\penalty0 324--345, 1952.
\newblock ISSN 00063444.
\newblock URL \url{http://www.jstor.org/stable/2334029}.

\bibitem[Chauhan et~al.(2023)Chauhan, Wang, and
  Wang]{chauhan-etal-2023-learning}
Jatin Chauhan, Xiaoxuan Wang, and Wei Wang.
\newblock Learning under label proportions for text classification.
\newblock In \emph{Findings of the Association for Computational Linguistics:
  EMNLP 2023}, pages 12210--12223, Singapore, 2023. Association for
  Computational Linguistics.
\newblock \doi{10.18653/v1/2023.findings-emnlp.817}.
\newblock URL \url{https://aclanthology.org/2023.findings-emnlp.817}.

\bibitem[Chen et~al.(2004)Chen, Huang, and Ramakrishnan]{CHR}
L.~Chen, Z.~Huang, and R.~Ramakrishnan.
\newblock Cost-based labeling of groups of mass spectra.
\newblock In \emph{Proc. ACM SIGMOD International Conference on Management of
  Data}, pages 167--178, 2004.

\bibitem[de~Freitas and K{\"{u}}ck(2005)]{FK05}
N.~de~Freitas and H.~K{\"{u}}ck.
\newblock Learning about individuals from group statistics.
\newblock In \emph{Proc. {UAI}}, pages 332--339, 2005.

\bibitem[Dery et~al.(2017)Dery, Nachman, Rubbo, and Schwartzman]{DNRS}
L.~M. Dery, B.~Nachman, F.~Rubbo, and A.~Schwartzman.
\newblock Weakly supervised classification in high energy physics.
\newblock \emph{Journal of High Energy Physics}, 2017\penalty0 (5):\penalty0
  1--11, 2017.

\bibitem[Dietterich et~al.(1997)Dietterich, Lathrop, and
  Lozano{-}P{\'{e}}rez]{DLL97}
Thomas~G. Dietterich, Richard~H. Lathrop, and Tom{\'{a}}s Lozano{-}P{\'{e}}rez.
\newblock Solving the multiple instance problem with axis-parallel rectangles.
\newblock \emph{Artif. Intell.}, 89\penalty0 (1-2):\penalty0 31--71, 1997.

\bibitem[Hendrycks et~al.(2021)Hendrycks, Burns, Kadavath, Arora, Basart, Tang,
  Song, and Steinhardt]{hendrycks2021math}
Dan Hendrycks, Collin Burns, Saurav Kadavath, Akul Arora, Steven Basart, Eric
  Tang, Dawn Song, and Jacob Steinhardt.
\newblock Measuring mathematical problem solving with the math dataset, 2021.

\bibitem[Hofst{\"a}tter et~al.(2020)Hofst{\"a}tter, Zlabinger, Sertkan,
  Schr{\"o}der, and Hanbury]{Hofstaetter2020_fira}
Sebastian Hofst{\"a}tter, Markus Zlabinger, Mete Sertkan, Michael Schr{\"o}der,
  and Allan Hanbury.
\newblock Fine-grained relevance annotations for multi-task document ranking
  and question answering.
\newblock In \emph{Proc. of CIKM}, 2020.

\bibitem[Honovich et~al.(2022)Honovich, Aharoni, Herzig, Taitelbaum, Kukliansy,
  Cohen, Scialom, Szpektor, Hassidim, and Matias]{Honovich2022TRUERF}
Or~Honovich, Roee Aharoni, Jonathan Herzig, Hagai Taitelbaum, Doron Kukliansy,
  Vered Cohen, Thomas Scialom, Idan Szpektor, Avinatan Hassidim, and Yossi
  Matias.
\newblock True: Re-evaluating factual consistency evaluation.
\newblock In \emph{Workshop on Document-grounded Dialogue and Conversational
  Question Answering}, 2022.
\newblock URL \url{https://api.semanticscholar.org/CorpusID:247694170}.

\bibitem[Ilse et~al.(2018)Ilse, Tomczak, and Welling]{ilse2018attention}
Maximilian Ilse, Jakub Tomczak, and Max Welling.
\newblock Attention-based deep multiple instance learning.
\newblock In \emph{International conference on machine learning}, pages
  2127--2136. PMLR, 2018.

\bibitem[Kotzias et~al.(2015)Kotzias, Denil, de~Freitas, and Smyth]{KDFS15}
D.~Kotzias, M.~Denil, N.~de~Freitas, and P.~Smyth.
\newblock From group to individual labels using deep features.
\newblock In \emph{Proc. SIGKDD}, pages 597--606, 2015.

\bibitem[Kulkarni et~al.(2020)Kulkarni, Chammas, Zhu, Sha, and
  Ie]{kulkarni2020aquamuse}
Sayali Kulkarni, Sheide Chammas, Wan Zhu, Fei Sha, and Eugene Ie.
\newblock Aquamuse: Automatically generating datasets for query-based
  multi-document summarization, 2020.

\bibitem[Li et~al.(2022)Li, Tomko, Vasardani, and
  Baldwin]{li-etal-2022-multispanqa}
Haonan Li, Martin Tomko, Maria Vasardani, and Timothy Baldwin.
\newblock {M}ulti{S}pan{QA}: A dataset for multi-span question answering.
\newblock In Marine Carpuat, Marie-Catherine de~Marneffe, and Ivan~Vladimir
  Meza~Ruiz, editors, \emph{Proceedings of the 2022 Conference of the North
  American Chapter of the Association for Computational Linguistics: Human
  Language Technologies}, pages 1250--1260, Seattle, United States, July 2022.
  Association for Computational Linguistics.
\newblock \doi{10.18653/v1/2022.naacl-main.90}.
\newblock URL \url{https://aclanthology.org/2022.naacl-main.90}.

\bibitem[Lightman et~al.(2023)Lightman, Kosaraju, Burda, Edwards, Baker, Lee,
  Leike, Schulman, Sutskever, and Cobbe]{lightman2023lets}
Hunter Lightman, Vineet Kosaraju, Yura Burda, Harri Edwards, Bowen Baker, Teddy
  Lee, Jan Leike, John Schulman, Ilya Sutskever, and Karl Cobbe.
\newblock Let's verify step by step, 2023.

\bibitem[Liu et~al.(2019)Liu, Wang, Qi, Tian, and Shi]{LWQTS19}
J.~Liu, B.~Wang, Z.~Qi, Y.~Tian, and Y.~Shi.
\newblock Learning from label proportions with generative adversarial networks.
\newblock In \emph{Proc. {NeurIPS}}, pages 7167--7177, 2019.

\bibitem[Liu et~al.(2021)Liu, Wang, Shen, Qi, and Tian]{LWSQT21}
Jiabin Liu, Bo~Wang, Xin Shen, Zhiquan Qi, and Yingjie Tian.
\newblock Two-stage training for learning from label proportions.
\newblock In Zhi-Hua Zhou, editor, \emph{Proceedings of the Thirtieth
  International Joint Conference on Artificial Intelligence, {IJCAI-21}}, pages
  2737--2743. International Joint Conferences on Artificial Intelligence
  Organization, 8 2021.
\newblock \doi{10.24963/ijcai.2021/377}.
\newblock URL \url{https://doi.org/10.24963/ijcai.2021/377}.
\newblock Main Track.

\bibitem[Liu et~al.(2022)Liu, Kong, Huang, Mao, and
  Xue]{liu-etal-2022-multiple}
Jiexi Liu, Dehan Kong, Longtao Huang, Dinghui Mao, and Hui Xue.
\newblock Multiple instance learning for offensive language detection.
\newblock In \emph{Findings of the Association for Computational Linguistics:
  EMNLP 2022}, pages 7387--7396. Association for Computational Linguistics,
  2022.
\newblock \doi{10.18653/v1/2022.findings-emnlp.546}.
\newblock URL \url{https://aclanthology.org/2022.findings-emnlp.546}.

\bibitem[Lozano-Pérez and Yang(2000)]{LY00}
T.~Lozano-Pérez and C.~Yang.
\newblock Image database retrieval with multiple-instance learning techniques.
\newblock In \emph{Proc. ICDE}, page 233, 2000.

\bibitem[Luo et~al.(2020)Luo, Guillory, Shi, Ke, Wan, Darrell, and
  Xu]{LGSKWDX20}
Zhekun Luo, Devin Guillory, Baifeng Shi, Wei Ke, Fang Wan, Trevor Darrell, and
  Huijuan Xu.
\newblock Weakly-supervised action localization with expectation-maximization
  multi-instance learning.
\newblock In \emph{Computer Vision -- ECCV 2020}, pages 729--745, Cham, 2020.
  Springer International Publishing.

\bibitem[Maron(1998)]{M98}
O.~Maron.
\newblock \emph{Learning from ambiguity}.
\newblock PhD thesis, Massassachusetts Institute of Technology, 1998.

\bibitem[Maron and Lozano-P\'{e}rez(1997)]{ML97}
Oded Maron and Tom\'{a}s Lozano-P\'{e}rez.
\newblock A framework for multiple-instance learning.
\newblock NIPS'97, page 570–576, 1997.

\bibitem[Musicant et~al.(2007)Musicant, Christensen, and Olson]{MCO07}
D.~R. Musicant, J.~M. Christensen, and J.~F. Olson.
\newblock Supervised learning by training on aggregate outputs.
\newblock In \emph{Proc. {ICDM}}, pages 252--261. {IEEE} Computer Society,
  2007.

\bibitem[Pappas and Popescu-Belis(2014)]{pappas-popescu-belis-2014-explaining}
Nikolaos Pappas and Andrei Popescu-Belis.
\newblock Explaining the stars: Weighted multiple-instance learning for
  aspect-based sentiment analysis.
\newblock In \emph{Proceedings of the 2014 Conference on Empirical Methods in
  Natural Language Processing ({EMNLP})}, pages 455--466, Doha, Qatar, October
  2014. Association for Computational Linguistics.
\newblock \doi{10.3115/v1/D14-1052}.
\newblock URL \url{https://aclanthology.org/D14-1052}.

\bibitem[Pappas and Popescu-Belis(2017)]{pappas2017explicit}
Nikolaos Pappas and Andrei Popescu-Belis.
\newblock Explicit document modeling through weighted multiple-instance
  learning.
\newblock \emph{Journal of Artificial Intelligence Research}, 58:\penalty0
  591--626, 2017.

\bibitem[Patrini et~al.(2014)Patrini, Nock, Caetano, and Rivera]{PNCR14}
G.~Patrini, R.~Nock, T.~S. Caetano, and P.~Rivera.
\newblock (almost) no label no cry.
\newblock In \emph{Proc. Advances in Neural Information Processing Systems},
  pages 190--198, 2014.

\bibitem[Perez-Beltrachini et~al.(2019)Perez-Beltrachini, Liu, and
  Lapata]{perez-beltrachini-etal-2019-wikicatsum}
Laura Perez-Beltrachini, Yang Liu, and Mirella Lapata.
\newblock Generating summaries with topic templates and structured
  convolutional decoders.
\newblock In Anna Korhonen, David Traum, and Llu{\'\i}s M{\`a}rquez, editors,
  \emph{Proceedings of the 57th Annual Meeting of the Association for
  Computational Linguistics}, pages 5107--5116, Florence, Italy, July 2019.
  Association for Computational Linguistics.
\newblock \doi{10.18653/v1/P19-1504}.
\newblock URL \url{https://aclanthology.org/P19-1504}.

\bibitem[Quadrianto et~al.(2009)Quadrianto, Smola, Caetano, and Le]{QSCL09}
N.~Quadrianto, A.~J. Smola, T.~S. Caetano, and Q.~V. Le.
\newblock Estimating labels from label proportions.
\newblock \emph{J. Mach. Learn. Res.}, 10:\penalty0 2349--2374, 2009.

\bibitem[Raffel et~al.(2023)Raffel, Shazeer, Roberts, Lee, Narang, Matena,
  Zhou, Li, and Liu]{raffel2023exploring}
Colin Raffel, Noam Shazeer, Adam Roberts, Katherine Lee, Sharan Narang, Michael
  Matena, Yanqi Zhou, Wei Li, and Peter~J. Liu.
\newblock Exploring the limits of transfer learning with a unified text-to-text
  transformer, 2023.

\bibitem[Ramon and De~Raedt(2000)]{ramon2000multi}
Jan Ramon and Luc De~Raedt.
\newblock Multi instance neural networks.
\newblock In \emph{Proceedings of the ICML-2000 workshop on attribute-value and
  relational learning}, pages 53--60, 2000.

\bibitem[Ray and Craven(2005)]{RC05}
Soumya Ray and Mark Craven.
\newblock Supervised versus multiple instance learning: an empirical
  comparison.
\newblock In \emph{Proc. ICML}, page 697–704, 2005.

\bibitem[Rueping(2010)]{R10}
S.~Rueping.
\newblock {SVM} classifier estimation from group probabilities.
\newblock In \emph{Proc. ICML}, pages 911--918, 2010.

\bibitem[Saket et~al.(2022)Saket, Raghuveer, and Ravindran]{SRR}
Rishi Saket, Aravindan Raghuveer, and Balaraman Ravindran.
\newblock On combining bags to better learn from label proportions.
\newblock In \emph{{AISTATS}}, volume 151 of \emph{Proceedings of Machine
  Learning Research}, pages 5913--5927. {PMLR}, 2022.
\newblock URL \url{https://proceedings.mlr.press/v151/saket22a.html}.

\bibitem[Scott and Zhang(2020)]{SZ20}
C.~Scott and J.~Zhang.
\newblock Learning from label proportions: {A} mutual contamination framework.
\newblock In \emph{Proc. NeurIPS}, 2020.

\bibitem[Sikka et~al.(2013)Sikka, Dhall, and Bartlett]{SDB13}
Karan Sikka, Abhinav Dhall, and Marian Bartlett.
\newblock Weakly supervised pain localization using multiple instance learning.
\newblock In \emph{2013 10th {IEEE} International Conference and Workshops on
  Automatic Face and Gesture Recognition (FG)}, pages 1--8, 2013.

\bibitem[Stelmakh et~al.(2023)Stelmakh, Luan, Dhingra, and
  Chang]{stelmakh2023asqa}
Ivan Stelmakh, Yi~Luan, Bhuwan Dhingra, and Ming-Wei Chang.
\newblock Asqa: Factoid questions meet long-form answers, 2023.

\bibitem[Wang et~al.(2023)Wang, Tang, Chen, Cheung, and Heng]{WFCCH23}
Xi~Wang, Fangyao Tang, Hao Chen, Carol~Y. Cheung, and Pheng-Ann Heng.
\newblock Deep semi-supervised multiple instance learning with self-correction
  for dme classification from oct images.
\newblock \emph{Medical Image Analysis}, 83:\penalty0 102673, 2023.
\newblock ISSN 1361-8415.
\newblock \doi{https://doi.org/10.1016/j.media.2022.102673}.
\newblock URL
  \url{https://www.sciencedirect.com/science/article/pii/S1361841522003012}.

\bibitem[Wu et~al.(2015{\natexlab{a}})Wu, Yu, Huang, and Yu]{WYY15}
J.~Wu, Yinan Yu, Chang Huang, and Kai Yu.
\newblock Deep multiple instance learning for image classification and
  auto-annotation.
\newblock In \emph{Proc. {CVPR}}, pages 3460--3469, 2015{\natexlab{a}}.

\bibitem[Wu et~al.(2015{\natexlab{b}})Wu, Yu, Huang, and Yu]{wu2015deep}
Jiajun Wu, Yinan Yu, Chang Huang, and Kai Yu.
\newblock Deep multiple instance learning for image classification and
  auto-annotation.
\newblock In \emph{Proceedings of the IEEE conference on computer vision and
  pattern recognition}, pages 3460--3469, 2015{\natexlab{b}}.

\bibitem[Wu et~al.(2023)Wu, Hu, Shi, Dziri, Suhr, Ammanabrolu, Smith,
  Ostendorf, and Hajishirzi]{wu2023finegrained}
Zeqiu Wu, Yushi Hu, Weijia Shi, Nouha Dziri, Alane Suhr, Prithviraj
  Ammanabrolu, Noah~A. Smith, Mari Ostendorf, and Hannaneh Hajishirzi.
\newblock Fine-grained human feedback gives better rewards for language model
  training, 2023.

\bibitem[Yin et~al.(2021)Yin, Radev, and Xiong]{Yin_2021docnli}
Wenpeng Yin, Dragomir Radev, and Caiming Xiong.
\newblock Docnli: A large-scale dataset for document-level natural language
  inference.
\newblock In \emph{Findings of the Association for Computational Linguistics:
  ACL-IJCNLP 2021}. Association for Computational Linguistics, 2021.
\newblock \doi{10.18653/v1/2021.findings-acl.435}.
\newblock URL \url{http://dx.doi.org/10.18653/v1/2021.findings-acl.435}.

\bibitem[Yu et~al.(2013)Yu, Liu, Kumar, Jebara, and Chang]{YLKJC13}
F.~X. Yu, D.~Liu, S.~Kumar, T.~Jebara, and S.~F. Chang.
\newblock $\propto${SVM} for learning with label proportions.
\newblock In \emph{Proc. ICML}, volume~28, pages 504--512, 2013.

\bibitem[Zhang et~al.(2005)Zhang, Platt, and Viola]{ZPV05}
Cha Zhang, John Platt, and Paul Viola.
\newblock Multiple instance boosting for object detection.
\newblock In \emph{Advances in Neural Information Processing Systems},
  volume~18. MIT Press, 2005.

\bibitem[Zhang and Goldman(2001)]{ZG01}
Qi~Zhang and Sally Goldman.
\newblock Em-dd: An improved multiple-instance learning technique.
\newblock \emph{Advances in neural information processing systems}, 14, 2001.

\end{thebibliography}

\appendix

\section{Task Specific Discussion of Experimental Results}\label{sec:dataset_results_discussion}
Following is the detailed per-task analysis:

\smallskip
\noindent
\textbf{Long-form Question Answering.} As presented in Table \ref{tab:all_except_fira_results_test_split}, the model trained using our PriorsBagLoss method on preference labels with the cosine-similarity prior has the best AUC-ROC, AUC-PR and accuracy scores in the experiments on the QA-feedback dataset. It outperforms both BagLoss as well as the cosine-similarity based baselines, the latter one by a significant margin.  However, we observe that the performance of the correlation prior based variant is worse than that of the BagLoss baseline which itself is worse than the Response-level trained baseline.  

\smallskip
\noindent
\textbf{Retrieval.} In the MultiSpan-QA dataset experiments (Table \ref{tab:all_except_fira_results_test_split}) we observe that the model trained by applying PsLab on the predictions of the model trained on PriorsBagLoss with a combination of the cosine-similarity and correlation priors achieves the best AUC-ROC and AUC-PR scores (by a significant gap) among the bag level baselines, while the variant with only the cosine-similarity prior performs the second bet on these metrics. However, the Response-level trained model and BagLoss achieve marginally higher accuracy scores. All these methods also handily outperform the cosine-similarity baseline. From the FiRA dataset results (Table \ref{tab:fira_results_test_split}) we observe that the our prior augmented BagLoss method, specifically the using both the priors, performs the best on mae as well as mse metrics. 

\smallskip
\noindent
\textbf{Summarization.} The experimental results on the WikiCatSum dataset, presented in Table \ref{tab:all_except_fira_results_test_split}, show that our PriorsBagLoss method with different combinations of the cosine-similarity and the correlation priors, or using NLI as a prior, as well as the PsLab method on these models yield the best performance (by a significant margin) in terms of AUC-ROC, AUC-PR and accuracy metrics. The comparative baselines are the BagLoss, NLI and cosine-similarity. A similar trend is observed on the AquaMuse dataset (Table \ref{tab:all_except_fira_results_test_split}) on which the our methods significantly outperform the bag-level baselines, in particular the PsLab applied to the PriorsBagLoss method yields the best performing model.

\smallskip
\noindent
\textbf{Math Reasoning.} On the PRM800k dataset, we observe from the experimental evaluations (Table \ref{tab:all_except_fira_results_test_split}) that the BagLoss and PriorsBagLoss methods are best performing among the bag-level baselines and also outperform the cosine-similarity baselines. While PriorsBagLoss using a combination of cosine-similarity and correlation priors achieves better AUC-ROC scores, BagLoss has significantly better accuracy while the AUC-PR scores are similar.

\section{Results for Differentiable Minimum Approximations} \label{sec:minappox}
As mentioned in Sec. \ref{sec:our_techniques}, in the binary-label case, the standard baseline is ${\sf Mult}$ which is just the product of the probabilities. 
More sophisticated approximations that we include in our study  are  ${\sf LSE}$~\cite{ramon2000multi}, ${\sf ISR}$, ${\sf NOR}$ and ${\sf GM}$~\cite{ZPV05} (see Sec. 2.4.1) of \cite{babenko2008multiple} for details). 
We employ the in built TensorFlow (TF) approximation ${\sf tf\_reduce\_min}$ in our experiments,  noting that ${\sf MAX}$ can be derived from ${\sf MIN}$ applied to flipped variables in the binary case. For integer labels we use the respective TF approximations for ${\sf MIN}$ and ${\sf MAX}$. The gradient of ${\sf tf\_reduce\_min}$ over a list of variables is non-zero only for those variables whose value is at the minimum.  

Table \ref{app_tab:min_approx_wikicatsum} has an ablation of Mult, GM and ${\sf tf\_reduce\_min}$ for the BagLoss and PriorBagLoss methods on the WikiCatSum dataset, demonstrating that ${\sf tf\_reduce\_min}$ outperforms the others in AUC-ROC and AUC-PR metrics.

\begin{table}[]
\centering
\begin{adjustbox}{width = 0.5\linewidth}
\begin{tabular}{@{}llrr@{}}
\toprule
\textbf{Model} & \textbf{AND Approx} & \multicolumn{1}{l}{\textbf{AUC-ROC}} & \multicolumn{1}{l}{\textbf{AUC-PR}} \\ \midrule
Instance Baseline                       & -              & 0.837 & 0.894 \\
\midrule
\multirow{3}{*}{BagLoss}                & Mult        & 0.449 & 0.785 \\
                                        & GM             & 0.463 & 0.824 \\
                                        & tf.reduce\_min & 0.478 & 0.829 \\
                                        \midrule
\multirow{3}{*}{PriorBagLoss(0.2, 0.1)} & Mult        & 0.599 & 0.858 \\
                                        & GM             & 0.631 & 0.862 \\
                                        & tf.reduce\_min & 0.643 & 0.877 \\ 
\bottomrule
\end{tabular}
\end{adjustbox}
\caption{Results for differentiable AND approximations on WikiCatSum dataset}

\label{app_tab:min_approx_wikicatsum}

\end{table}

\section{Aggregate and instance-level evaluations}
We also include evaluations of the various methods on a test set of bags w.r.t. bag-level metrics using the corresponding ${\sf AGG}$ approximations. 
Tables \ref{app_tab:multispan_agg_inst_results_test_split}, \ref{app_tab:qa_pref_results_test_split}, \ref{app_tab:fira_results_test_split}, \ref{app_tab:wikicatsum_results_test_split}, \ref{app_tab:prm800k_results_test_split} and \ref{app_tab:qa_pref_results_test_split} contain the aggregate as well as instance evaluations.
\begin{table*}[htb]
\centering
    \begin{adjustbox}{width=\linewidth}
        \begin{tabular}{llrrrrr}
        \toprule
        \textbf{Evaluation} & \textbf{Model} & \multicolumn{1}{l}{\textbf{AUC-ROC}} & \multicolumn{1}{l}{\textbf{AUC-PR}} & \textbf{Accuracy} & \textbf{Precision} & \textbf{Recall} \\
        \midrule
        \multirow{7}{*}{Aggregate} & Cosine Similarity & 0.488 & 0.287 & \multicolumn{1}{r}{0.698} & \multicolumn{1}{r}{0} & \multicolumn{1}{r}{0} \\
         & NLI & 0.6855 & 0.522 & \multicolumn{1}{r}{0.7453} & \multicolumn{1}{r}{0.7883} & \multicolumn{1}{r}{0.319} \\
         & Response-level Model & \multicolumn{1}{l}{0.681 ± 0.012} & \multicolumn{1}{l}{0.525 ± 0.081} & 0.726 ± 0.033 & 0.752 ± 0.011 & 0.428 ± 0.063 \\
         & Sentence-level Model & 0.653 ± 0.076 & 0.462 ± 0.050 & 0.723 ± 0.015 & 0.599 ± 0.093 & 0.435 ± 0.187 \\
         & BagLoss & 0.678 ± 0.082 & 0.525 ± 0.072 & 0.718 ± 0.057 & 0.717 ± 0.014 & 0.391 ± 0.033 \\
         & 0.8 BagLoss + 0.2 P1 & 0.683 ± 0.053 & 0.527 ± 0.02 & 0.722 ± 0.035 & 0.748 ± 0.009 & 0.461 ± 0.01 \\
         & 0.7 BagLoss + 0.2 P2 + 0.1 P1 & 0.665 ± 0.061 & 0.491 ± 0.04 & 0.727 ± 0.029 & 0.693 ± 0.018 & 0.316 ± 0.037 \\
         \midrule
        \multirow{8}{*}{Instance} & Cosine Similarity & 0.455 & 0.135 & \multicolumn{1}{r}{0.851} & \multicolumn{1}{r}{0} & \multicolumn{1}{r}{0} \\
         & NLI & 0.631 & 0.366 & \multicolumn{1}{r}{0.859} & \multicolumn{1}{r}{0.872} & \multicolumn{1}{r}{0.178} \\
         & Response-level Model & 0.583 ± 0.187 & 0.217 ± 0.094 & 0.852 ± 0.003 & 0.529 ± 0.074 & 0.086 ± 0.04 \\
         & Sentence-level Model & 0.729 ± 0.016 & 0.354 ± 0.008 & 0.861 ± 0.046 & 0.717 ± 0.011 & 0.438 ± 0.072 \\
         & BagLoss & 0.661 ± 0.092 & 0.309 ± 0.127 & 0.852 ± 0.133 & 0.711 ± 0.074 & 0.24 ± 0.188 \\
         & 0.8 BagLoss + 0.2 P1 & 0.669 ± 0.063 & 0.311 ± 0.059 & 0.838 ± 0.071 & 0.65 ± 0.089 & 0.189 ± 0.112 \\
         & 0.7 BagLoss + 0.2 P2 + 0.1 P1 & 0.625 ± 0.07 & 0.271 ± 0.039 & 0.851 ± 0.021 & 0.639 ± 0.189 & 0.135 ± 0.098 \\
         & \fgl & \multicolumn{1}{l}{0.693 ± 0.115} & \multicolumn{1}{l}{0.326 ± 0.071} & 0.842 ± 0.052 & 0.676 ± 0.043 & 0.228 ± 0.091 \\

        \bottomrule
        \end{tabular}
\end{adjustbox}
\caption{Comparison of aggregate and instance-level performance on MultiSpanQA Dataset}
\label{app_tab:multispan_agg_inst_results_test_split}
\end{table*}

\begin{table*}[htb]
\centering
\begin{adjustbox}{width=0.8\linewidth}

\begin{tabular}{@{}llrrrrr@{}}
\toprule
\textbf{Evaluation} &
  \textbf{Method} &
  \multicolumn{1}{l}{\textbf{AUC-ROC}} &
  \multicolumn{1}{l}{\textbf{AUC-PR}} &
  \multicolumn{1}{l}{\textbf{Accuracy}} &
  \multicolumn{1}{l}{\textbf{Precision}} &
  \multicolumn{1}{l}{\textbf{Recall}} \\
  \midrule
\textbf{Preference} & Cosine Similarity    & 0.4978 & 0.3952 & 0.477  & 0.379 & 0.4985 \\
\textbf{}           & Response-level Model & 0.546  & 0.4651 & 0.553  & 0.437 & 0.539  \\
\textbf{}           & BagLoss              & 0.543  & 0.4644 & 0.5463 & 0.442 & 0.5324 \\
\textbf{}           & PriorBagLoss(0.2,0)  & 0.568  & 0.4658 & 0.574  & 0.439 & 0.5467 \\
\midrule
\textbf{Instance}   & Sentence-level Model & 0.647  & 0.611  & 0.686  & 0.722 & 0.418  \\
                    & Cosine Similarity    & 0.535 & 0.526  & 0.483  & 0.893 & 0.134  \\
                    & Response-level Model & 0.491 & 0.4643 & 0.453  & 0.882 & 0.278      \\
                    & BagLoss              & 0.509 & 0.5269 & 0.5167 & 0.814 & 0.36    \\
                    & PriorBagLoss(0.2, 0) & 0.516 & 0.4936 & 0.508  & 0.647 & 0.715 \\
                    \bottomrule
\end{tabular}

\end{adjustbox}
\caption{Comparison of Preference and instance-level evaluation on QA Preference Feedback Dataset}
\label{app_tab:qa_pref_results_test_split}
\end{table*}

\begin{table*}[htb]
\centering
\begin{adjustbox}{width=0.6\linewidth}
\begin{tabular}{llcc}
\toprule
\textbf{Evaluation} & \textbf{Model} & \multicolumn{1}{l}{\textbf{MAE}} & \multicolumn{1}{l}{\textbf{MSE}} \\
\midrule
\multirow{3}{*}{Aggregate} & Instance-level Model & 0.375 ± 0.09 & 0.247 ± 0.113 \\
 & Response-level Model & 0.320 ± 0.042 & 0.197 ± 0.017\\
 & BagLoss & 0.326 ± 0.021 & 0.209 ± 0.007 \\
 \midrule
\multirow{3}{*}{Instance} & Instance-level Model & 0.283 ± 0.072 & 0.141 ± 0.088 \\
 & Response-level Model & 0.319 ± 0.047 & 0.186 ± 0.098 \\
 & BagLoss & 0.304 ± 0.007 & 0.163 ± 0.002\\
 & PriorBagLoss(0.3, 0) & 0.298 ± 0.002 & 0.157 ± 0.004\\
 & PriorBagLoss(0.2, 0.2) & 0.294 ± 0.003 & 0.155 ± 0.001\\
 \bottomrule
\end{tabular}
\end{adjustbox}
\caption{Comparison of Aggregate and Instance-level evaluations on FirA Dataset}
\label{app_tab:fira_results_test_split}
\end{table*}

\begin{table*}[htb]
\centering
\begin{adjustbox}{width=0.8\linewidth}
\begin{tabular}{lccccc}
\toprule
\textbf{Method} & \textbf{AUC-ROC} & \textbf{AUC-PR} & \textbf{Accuracy} & \textbf{Precision} & \textbf{Recall} \\
\midrule
Instance Baseline & 0.837 ± 0.062 & 0.894 ± 0.085 & 0.718 ± 0.085 & 0.926 ± 0.001 & 0.733 ± 0.003 \\
NLI & \multicolumn{1}{r}{0.639} & \multicolumn{1}{r}{0.817} & \multicolumn{1}{r}{0.648} & \multicolumn{1}{r}{0.834} & \multicolumn{1}{r}{0.559} \\
Cosine Similarity & \multicolumn{1}{r}{0.408} & \multicolumn{1}{r}{0.829} & \multicolumn{1}{r}{0.362} & \multicolumn{1}{r}{0.719} & \multicolumn{1}{r}{0.276} \\
BagLoss & 0.477 ± 0.093 & 0.831 ± 0.052 & 0.562 ± 0.047 & 0.769 ± 0.018 & 0.319 ± 0.048 \\
PriorBagLoss(0.2, 0.1) & 0.641 ± 0.028 & 0.879 ± 0.013 & 0.651 ± 0.017 & 0.897 ± 0.009 & 0.658 ± 0.082 \\
\psl & 0.645 ± 0.038 & 0.879 ± 0.057 & 0.663 ± 0.091 & 0.884 ± 0.076 & 0.661 ± 0.092 \\
0.6*BagLoss + 0.4*NLI & \multicolumn{1}{r}{0.642} & \multicolumn{1}{r}{0.885} & \multicolumn{1}{r}{0.653} & \multicolumn{1}{r}{0.914} & \multicolumn{1}{r}{0.619} \\
\bottomrule
\end{tabular}
\end{adjustbox}
\caption{Instance-level Evaluation on WikiCatSum}
\label{app_tab:wikicatsum_results_test_split}
\end{table*}

\begin{table*}[htb]
\centering
\begin{adjustbox}{width=0.8\linewidth}
\begin{tabular}{lccccc}
\toprule
\textbf{Method}            & \textbf{AUC-ROC} & \textbf{AUC-PR} & \textbf{Accuracy} & \textbf{Precision} & \textbf{Recall} \\
\midrule
Sentence-level Model & 0.613 ± 0.028 & 0.928 ± 0.021 & 0.709 ± 0.051 & 0.902 ± 0.018 & 0.993 ± 0.004 \\
Response-level Model & 0.528 ± 0.145 & 0.895 ± 0.037 & 0.521 ± 0.082 & 0.851 ± 0.055 & 0.577 ± 0.065 \\
Cosine Similarity    & 0.420         & 0.873         & 0.496         & 0.888         & 0.683         \\
BagLoss                 & 0.569 ± 0.019 & 0.924 ± 0.011 & 0.688 ± 0.071 & 0.904 ± 0.014 & 0.955 ± 0.048 \\
PriorBagLoss(0.5, 0)    & 0.582 ± 0.034 & 0.927 ± 0.009 & 0.534 ± 0.044 & 0.920 ± 0.014 & 0.606 ± 0.037 \\
PriorBagLoss(0, 0.1)    & 0.579 ± 0.052 & 0.925 ± 0.017 & 0.603 ± 0.038 & 0.908 ± 0.020 & 0.794 ± 0.055 \\
PriorBagLoss(0.1, 0.1) & 0.580 ± 0.068    & 0.926 ± 0.024   & 0.563 ± 0.049     & 0.914 ± 0.028      & 0.708 ± 0.077   \\
psl                  & 0.597 ± 0.093 & 0.927 ± 0.004 & 0.578 ± 0.063 & 0.911 ± 0.009 & 0.713 ± 0.128 \\ 
\bottomrule
\end{tabular}
\end{adjustbox}
\caption{Instance-level Evaluation on PRM800K}
\label{app_tab:prm800k_results_test_split}
\end{table*}

\begin{table*}[htb]
\centering
\begin{tabular}{@{}lrrrrr@{}}
\toprule
\textbf{Method} &
  \multicolumn{1}{l}{\textbf{AUC-ROC}} &
  \multicolumn{1}{l}{\textbf{AUC-PR}} &
  \multicolumn{1}{l}{\textbf{Accuracy}} &
  \multicolumn{1}{l}{\textbf{Precision}} &
  \multicolumn{1}{l}{\textbf{Recall}} \\
 \midrule
Sentence-level Model         & 0.648 & 0.611  & 0.686  & 0.722 & 0.418 \\
Cosine Similarity            & 0.535 & 0.526  & 0.483  & 0.893 & 0.134 \\
Response-level Model         & 0.491 & 0.4643 & 0.453  & 0.882 & 0.278 \\
BagLoss                         & 0.509 & 0.5269 & 0.5167 & 0.814 & 0.36  \\
PriorBagLoss(0.2, 0)    & 0.516 & 0.4936 & 0.508  & 0.647 & 0.715 \\
PriorBagLoss(0, 0.4)        & 0.527 & 0.515  & 0.529  & 0.519 & 0.763 \\
PriorBagLoss(0.2, 0.5)       & 0.532 & 0.522  & 0.521  & 0.738 & 0.469\\
\bottomrule
\end{tabular}
\caption{QA Preference Feedback Results (Instance Evaluation)}
\label{app_tab:qa_pref_results_test_split_instance}
\end{table*}

\section{Hyperparameter Tuning}\label{sec:hyper}

Table \ref{app_tab:hyperparam} contains the best weights for our prior augmented BagLoss method on different datasets, along with the best learning rates and batch size in the bag-level training.

\begin{table*}[htb]
\centering
\begin{tabular}{lrrrrr}
\toprule
\textbf{Dataset} &
  \multicolumn{1}{l}{\textbf{$\alpha_1$}} &
  \multicolumn{1}{l}{\textbf{$\alpha_2$}} &
  \multicolumn{1}{l}{\textbf{$\alpha_3$}} &
  \multicolumn{1}{l}{\textbf{Learning Rate}} &
  \multicolumn{1}{l}{\textbf{Batch Size}} \\
  \midrule
QA-Feedback & 0.3 & 0.2 & 0.5 & 1e-5 & 256  \\
MultiSpanQA & 0.8 & 0.2 & 0   & 1e-3 & 512  \\
WikiCatSum  & 0.7 & 0.2 & 0.1 & 1e-4 & 1024 \\
FiRA        & 0.6 & 0.2 & 0.2 & 1e-5 & 1024 \\
AquaMuSe    & 0.7 & 0.2 & 0.1 & 1e-3 & 512  \\
PRM800K     & 0.8 & 0.1 & 0.1 & 1e-4 & 64  \\
\bottomrule
\end{tabular}
\caption{$\alpha_1$, $\alpha_2$, and $\alpha_3$ represent the coefficients of the BagLoss, cosine similarity prior and correlation prior terms in the loss function.}
\label{app_tab:hyperparam}
\end{table*}

\section{Examples of perturbed summaries from the WikiCatSum and Aquamuse Datasets} \label{sec:example_summary}
Tables \ref{app_tab:aquamuse_summary_examples} and \ref{app_tab:wikicatsum_summary_examples} contain the entailed and non-entailed (perturbed) summaries for the Aquamuse and WikiCatSum datasets respectively.

\begin{table*}[]
    \begin{small}
    \centering
    \begin{adjustbox}{width=\linewidth}
    \begin{tabular}{p{2cm}|p{12cm}}
    \toprule
     \textbf{Type} & \textbf{Summary} \\
     \midrule
     Reference Summary & She also is the first ever woman in Indian History to be nominated as the Rajya Sabha member. She is considered the most important revivalist in the Indian classical dance form of Bharatanatyam from its original' sadhir' style, prevalent amongst the temple dancers, Devadasis, she also worked for the re-establishment of traditional Indian arts and crafts.\\
     \midrule
     Word Replacement & She also is the first ever woman in Indian History to be nominated as the Rajya Sabha \textcolor{red}{Independent} member. She is considered the most important revivalist in the Indian classical dance form of \textcolor{red}{Kathak} from its Nautch 'sadhir' style, prevalent amongst the temple \textcolor{red}{singers}. Furthermore, she also advocated for the re-establishment of traditional Indian arts and crafts. \\
     \midrule
     Sentence Replacement & She also is the first ever woman in Indian History to be nominated as the Rajya Sabha member. She is considered the most important revivalist in the Indian classical dance form of Bharatanatyam from its original' sadhir' style, prevalent amongst the temple dancers. \textcolor{red}{She was also a strong advocate for animal welfare and environmental protection, actively participating in campaigns and legislative efforts throughout her life.}\\
    \bottomrule
    \end{tabular}
    \end{adjustbox}
    \caption{Example of the entailed and non-entailed versions of the summary from AquaMuSe Dataset. We either use entailed or non-entailed version.}
    \label{app_tab:aquamuse_summary_examples}
    \end{small}
\end{table*}

\begin{table*}[]
    \begin{small}
    \centering
    \begin{adjustbox}{width=\linewidth}
    \begin{tabular}{p{2cm}|p{12cm}}
    \toprule
     \textbf{Type} & \textbf{Summary} \\
     \midrule
     Reference Summary & the gold spangle ( autographa bractea ) is a moth of the family noctuidae . it is found in europe , across western siberia and the altai mountains , the northern caucasus , northern turkey and northern iran . its wingspan is 42 – 50 mm .  the forewings are brown and gray with large rhomboid golden marks .  the hindwings and body are lighter grayish brown .  the moth flies from july to august depending on the location , and migrates long distances . the larvae feed on a wide range of plants including hieracium , tussilago farfara , plantago , crepis paludosa , taraxacum , urtica , lamium , stachys and eupatorium cannabinum . \\
     \midrule
     Word Replacement & the gold spangle ( \textcolor{red}{autographa californica} ) is a moth of the family noctuidae . it is found in \textcolor{red}{western north america} , across \textcolor{red}{california} and the altai mountains , \textcolor{red}{south dakota} and \textcolor{red}{new mexico} . its wingspan is \textcolor{red}{16 - 25} mm . the forewings are \textcolor{red}{blue} and gray with \textcolor{red}{silver-white long lateral part} and a \textcolor{red}{patch of chestnut brown} . the hindwings and body are \textcolor{red}{a grayish tan} . the moth flies from \textcolor{red}{march to september} depending on the location , and migrates long distances . the larvae feed on a wide range of \textcolor{red}{herbaceous} plants including \textcolor{red}{legumes such as fabaceae , alfalfas , peas , taraxacum , urtica , lamium , stachys and eupatorium cannabinum} . \\
     \midrule
     Sentence Replacement & the gold spangle ( autographa bractea ) is a moth of the family noctuidae . it is found in europe , across western siberia and the altai mountains , the northern caucasus , northern turkey and northern iran . its wingspan is 42 – 50 mm . \textcolor{red}{the forewing has an inner line below middle finely golden in color, and the outer one is golden at the inner margin only} . the hindwings and body are lighter grayish brown . the moth flies from july to august depending on the location , and migrates long distances . \textcolor{red}{Occupying waste ground, gardens and moorland, this species is widespread and fairly common in the north of Britain} .\\
    \bottomrule
    \end{tabular}
    \end{adjustbox}
    \caption{Example of the entailed and non-entailed versions of the summary from WikiCatSum Dataset. We either use entailed or non-entailed version.}
    \label{app_tab:wikicatsum_summary_examples}
    \end{small}
\end{table*}

\end{document}